\documentclass[11pt]{article}

\usepackage[preprint]{acl}

\usepackage{times}
\usepackage{latexsym}

\usepackage{amsmath}
\usepackage[ruled,vlined,linesnumbered]{algorithm2e}
\SetKwInput{KwParam}{Hyperparameters}

\usepackage{booktabs}
\usepackage{multirow}
\usepackage{makecell}
\usepackage{array}
\usepackage{graphicx}
\usepackage[table]{xcolor}

\usepackage[T1]{fontenc}

\usepackage[utf8]{inputenc}

\usepackage{microtype}

\usepackage{inconsolata}

\usepackage{graphicx}
\usepackage{amsmath}
\usepackage{amssymb}

\usepackage{fancyhdr}
\fancypagestyle{wip}{%
  \fancyhf{}%
  \fancyhead[L]{Working in Progress}%
  \fancyfoot[C]{\thepage}%
}
\title{On-Policy Distillation with Curriculum Turn-level Guidance for Multi-turn Agents}

\author{
  \textbf{Gengsheng Li\textsuperscript{1,2}$^\ast$},
  \textbf{Mao Zheng\textsuperscript{3}$^\ast$},
  \textbf{Mingyang Song\textsuperscript{3}$^\ast$},
  \textbf{Ruiqi Liu\textsuperscript{2}},
\\
  \textbf{Tianyu Yang\textsuperscript{1,2}},
  \textbf{Jie Sun\textsuperscript{4}},
  \textbf{Qiyong Zhong\textsuperscript{5}},
  \textbf{Haiyun Guo\textsuperscript{1,2}},
\\
  \textbf{Junfeng Fang\textsuperscript{6}},
  \textbf{Dan Zhang\textsuperscript{6}},
  \textbf{Jinqiao Wang\textsuperscript{1,2,7}}
\\
\\
  \textsuperscript{1}Foundation Model Research Center, Institute of Automation, Chinese Academy of Sciences
\\
  \textsuperscript{2}School of Artificial Intelligence, University of Chinese Academy of Sciences
\\
  \textsuperscript{3}Large Language Model Department, Tencent,
  \textsuperscript{4}Shanghai Innovation Institute
\\
  \textsuperscript{5}Zhejiang University,
  \textsuperscript{6}National University of Singapore,
  \textsuperscript{7}Wuhan AI Research
\\
  \small{
    \texttt{ligengsheng2024@ia.ac.cn}
  }
}

\newcommand{\blfootnote}[1]{%
  \begingroup
  \renewcommand\thefootnote{}\footnote{#1}%
  \addtocounter{footnote}{-1}%
  \endgroup
}

\begin{document}
\maketitle
\thispagestyle{wip}
\blfootnote{$^\ast$Equal contribution. Code: \url{https://github.com/Zzzz-166/Guided-OPD}}
\begin{abstract}
Multi-turn agents that plan, invoke tools, and interact with environments offer a promising paradigm for solving complex tasks, yet their capabilities typically rely on very large models whose inference cost is prohibitive in practice.
On-Policy Distillation (OPD) is a natural recipe for transferring such capabilities to smaller students, but we find that it suffers a characteristic failure mode in this setting: small student errors compound across turns and push the trajectory out of the teacher's familiar state distribution, so the teacher's supervision becomes least reliable precisely where the student needs it most.
We propose Guided On-Policy Distillation (Guided-OPD), a simple yet effective algorithm that mixes teacher- and student-generated turns within each rollout and schedules the teacher's intervention probability along a curriculum that decays to zero.
Strong guidance keeps early trajectories close to the teacher distribution and is then gradually withdrawn to recover the purely on-policy regime used at inference.
On ALFWorld, ScienceWorld, and WebShop, distilling Qwen3 students from a Qwen3-30B-A3B teacher, Guided-OPD yields average relative gains of 21.1\% in Score and 25.5\% in Success Rate over vanilla OPD, with larger gains on smaller students.
\end{abstract}

\section{Introduction}
\label{sec:intro}

\begin{figure}[!t]
    \centering
    \includegraphics[width=\linewidth]{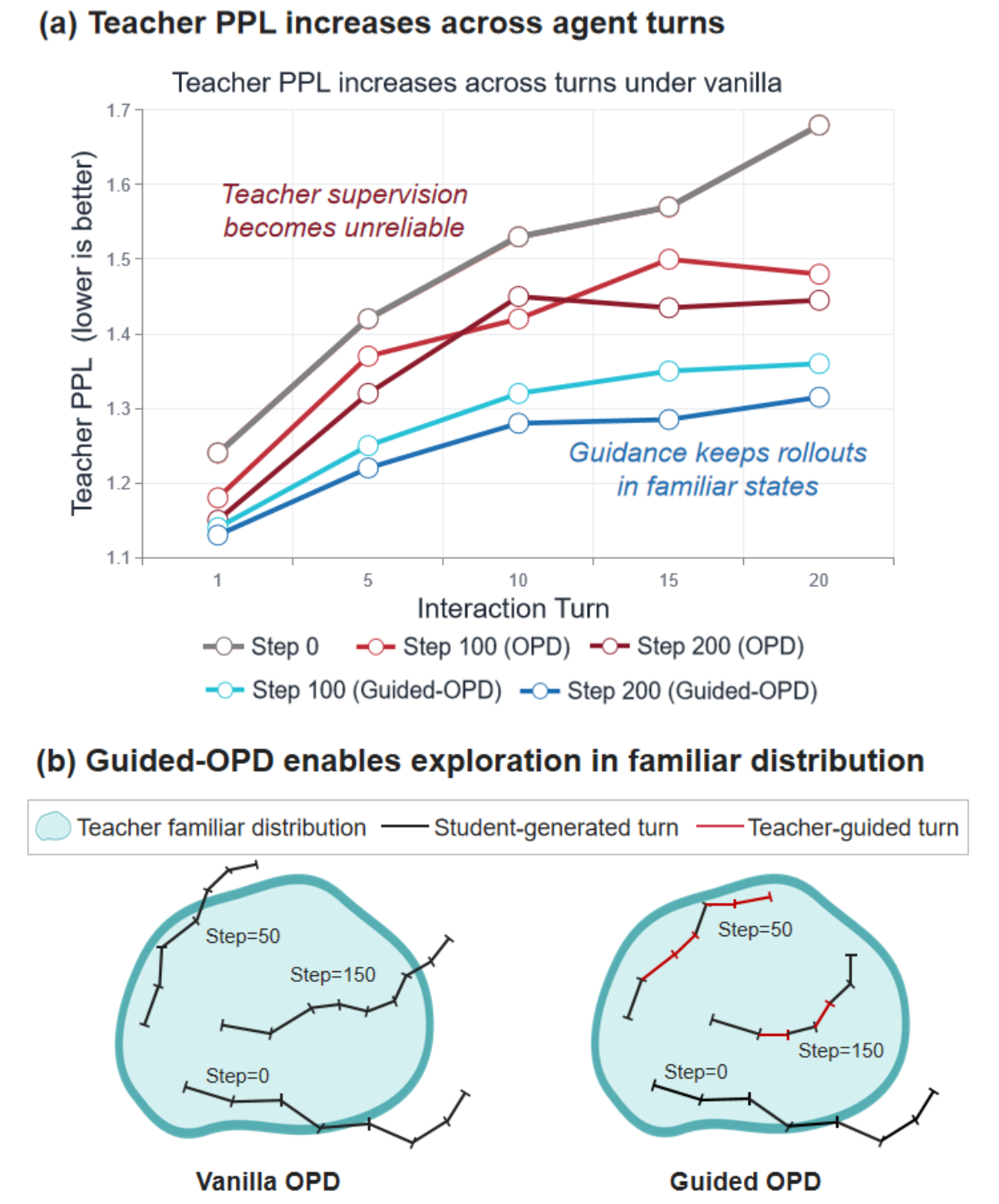}
    \caption{Failure mode of vanilla OPD on multi-turn agents and how Guided-OPD addresses it.
    (a) Teacher PPL on student-generated tokens versus interaction turn at different training steps: under vanilla OPD, PPL grows with the turn index and remains high even after $200$ steps, while Guided-OPD lowers PPL across all turns, with the gap widening at later turns.
    (b) Schematic of the rollout dynamics. The teal region denotes the teacher's familiar distribution, where black and red segments denote student- and teacher-generated turns. Vanilla OPD samples every turn from the student and quickly leaves this region, while Guided-OPD intersperses teacher turns to anchor the rollout, with the curriculum schedule injecting more teacher turns early and fewer as training proceeds.}
    \label{fig:teaser}
\end{figure}

Multi-turn agents that autonomously plan~\citep{wang2023plan}, use tools~\citep{schick2023toolformer,singh2025agentic}, and interact with their environment~\citep{react,chen2025reinforcement} are becoming an essential paradigm for solving complex tasks~\citep{xi2023rise,demy}. Such capabilities, however, are typically realized by models with very large parameter counts~\citep{qwen2025qwen3}, which incur substantial inference cost and deployment overhead~\citep{xu2024ondevice}. A central question for building practical agents is therefore how to stably transfer the strong agentic capabilities of a large teacher model to a much smaller student~\citep{xu2024kd_llm_survey,AgentDistillation,Agentdistill,sad,coa}. On-Policy Distillation (OPD) has recently emerged as an effective paradigm for transferring complex reasoning abilities~\citep{lu2025onpolicydistillation,mimov2flash,sun2026simct,zhong2026sod}. By aligning the teacher distribution directly on trajectories sampled from the student itself~\citep{opd,gu2024minillm}, 
OPD combines the training-inference consistency of on-policy learning with the dense supervision of token-level distillation, 
and yields strong results on reasoning tasks such as mathematics~\citep{eopd, tip} and code generation~\citep{uniopd, yang2026learning}. 
Nevertheless, previous work points out that OPD often suffers from noisy supervision signals~\citep{fu2026revisiting,li2026rethinking} and unstable optimization~\citep{manyfaces2026,stableopd2026,opdsurvey,srpo,sdrlvr}. Since multi-turn agents constitute a dynamic interactive setting that differs sharply from static reasoning~\citep{demy,zhou2024archer} and can plausibly amplify the issues already reported there~\citep{xue2025simpletir,skillsd}, whether OPD remains effective in this regime is an urgent open question.

In this paper, we apply OPD to multi-turn agents and identify a characteristic failure mode. 
For every agent turn, we compute the cross-entropy of the teacher with respect to the tokens generated by the student, and exponentiate it to obtain the teacher perplexity (PPL). 
As shown in Figure~\ref{fig:teaser}(a), under vanilla OPD the teacher PPL grows with the turn index, indicating that the reliability of the teacher's supervision degrades as the agent interacts more extensively with the environment. 
We attribute this behavior to a form of distribution drift that is intrinsic to the multi-turn setting. 
Within a single rollout, each student action triggers unpredictable environment feedback, and small biases introduced in early turns can propagate and compound across the trajectory, so that within only a handful of turns the student may steer the rollout out of the state distribution familiar to the teacher. 
On such trajectories the teacher itself becomes uncertain, and its output distribution no longer serves as a reliable distillation target. More critically, this effect is not naturally alleviated by longer training: as shown in Figure~\ref{fig:teaser}(a), the overall PPL drops only marginally from step 100 to step 200, suggesting that the elevated PPL early in training already constitutes a bottleneck that obstructs efficient knowledge transfer from the teacher to the student.

To address this issue, we propose Guided On-Policy Distillation (Guided-OPD), a simple yet effective training paradigm. Instead of letting the student complete an entire trajectory on its own, Guided-OPD samples each turn from the teacher with probability $\beta$ and from the student with probability $1-\beta$, with both models sharing the same context.
Building on this design, we further introduce a curriculum schedule that decays $\beta$ as training proceeds. Early in training, when the student is weak and prone to drift, a relatively large $\beta$ injects more teacher guidance and keeps the trajectory within the teacher's familiar distribution as much as possible, thereby securing reliable supervision. As the student improves and becomes less likely to push the rollout out of distribution, the decay of $\beta$ encourages broader self-exploration and gradually returns training to a purely on-policy regime, which avoids the training-inference mismatch.
As shown in Figure~\ref{fig:teaser}(a), Guided-OPD substantially lowers the teacher PPL and provides the student with more reliable supervision, while Figure~\ref{fig:teaser}(b) presents an intuitive schematic of the method.

We systematically validate Guided-OPD on three representative multi-turn agentic benchmarks: ALFWorld~\citep{ALFWorld}, ScienceWorld~\citep{ScienceWorld}, and WebShop~\citep{WebShop}, under three distillation settings that transfer a Qwen3-30B-A3B teacher to 0.6B, 1.7B, and 4B students. Across the three configurations Guided-OPD delivers average relative gains of $+21.1\%$ in Score and $+25.5\%$ in Success Rate over vanilla OPD, while reducing the average number of interaction turns by $6.0\%$. The improvement grows as the student becomes smaller: on the 0.6B student, the relative gains reach $+35.3\%$ in Score and $+32.8\%$ in Success Rate, with a $7.9\%$ reduction in interaction turns. 
Our main contributions are summarized as follows:
\begin{itemize}
    \item We identify a characteristic failure mode of OPD on multi-turn agents, in which the teacher PPL over student-generated tokens grows steadily along the trajectory and is not relieved by longer training, exposing rollout-level distribution drift as a fundamental obstacle to reliable distillation.
    \item We propose Guided-OPD, which mixes teacher- and student-sampled turns within each rollout and equips this mixture with a curriculum schedule, keeping early trajectories close to the teacher distribution while smoothly converging to a purely on-policy regime in later training stage.
    \item We validate Guided-OPD on ALFWorld, ScienceWorld, and WebShop across 0.6B, 1.7B, and 4B students distilled from Qwen3-30B-A3B, observing consistent gains in Score and Success Rate, fewer interaction turns, and larger benefits on smaller students. 
\end{itemize}

\section{Preliminary}
\label{sec:preliminary}

\textbf{Multi-turn Agent.}
We consider an agent interacting with an environment over a finite horizon $T$, indexed by $t \in \{0, \dots, T-1\}$.
At each turn $t$, the agent receives an observation $o_t$ and produces a response $a_t$ that consists of a reasoning trace followed by an executable action~\citep{react}, after which the environment returns the next observation $o_{t+1}$.
We use \emph{turn $t$} to refer to one such interaction round (i.e., the cycle $o_t \!\to\! a_t \!\to\! o_{t+1}$), and reserve $a_t$ for the agent-produced segment within it.
Since the environment is partially observable, we define the agent state as the full interaction history up to the current observation,
\begin{equation}
    h_t = (o_0, a_0, \dots, o_{t-1}, a_{t-1}, o_t),
    \label{eq:history}
\end{equation}
and a complete trajectory as $\tau = (h_0, a_0, \dots, h_{T-1}, a_{T-1})$, which terminates when a termination action is executed or when $T$ is reached.
The agent is parameterized as an autoregressive language model $\pi_\theta(a_t \mid h_t)$. 

\textbf{On-Policy Distillation.}
Given a frozen teacher policy $\pi_\phi$ and a student policy $\pi_\theta$, OPD aligns the student with the teacher \emph{under the student's own state distribution}: the student rolls out a trajectory $\tau \sim \pi_\theta$, and the teacher then scores $\tau$ via a token-level divergence at each visited state. The training objective is
\begin{equation}
\begin{aligned}
\mathcal{L}_{\mathrm{OPD}}(\theta)
&= \mathbb{E}_{\tau \sim \pi_\theta}
\Bigg[
\sum_{t=0}^{T-1}
\mathcal{D}\Big(
\pi_\phi(a_t \mid h_t), \\
&\qquad\qquad
\bar{\pi}_\theta(a_t \mid h_t)
\Big)
\Bigg].
\end{aligned}
\label{eq:opd}
\end{equation}
where $\mathcal{D}(\cdot,\cdot)$ is a token-level divergence between the teacher and student conditional distributions, generally including three standard instantiations:
\emph{(i) forward KL} $\mathcal{D}_{\mathrm{KL}}\!\big(\pi_\phi \,\Vert\, \pi_\theta\big)$, 
\emph{(ii) reverse KL} $\mathcal{D}_{\mathrm{KL}}\!\big(\pi_\theta \,\Vert\, \pi_\phi\big)$, 
and \emph{(iii) Jensen--Shannon divergence}, the symmetric interpolation $\tfrac{1}{2}\mathcal{D}_{\mathrm{KL}}(\pi_\phi \,\Vert\, M) + \tfrac{1}{2}\mathcal{D}_{\mathrm{KL}}(\pi_\theta \,\Vert\, M)$ with $M = \tfrac{1}{2}(\pi_\phi + \pi_\theta)$.

\section{Method}
\label{sec:method}

\begin{figure*}[t]
    \centering
    \includegraphics[width=\linewidth]{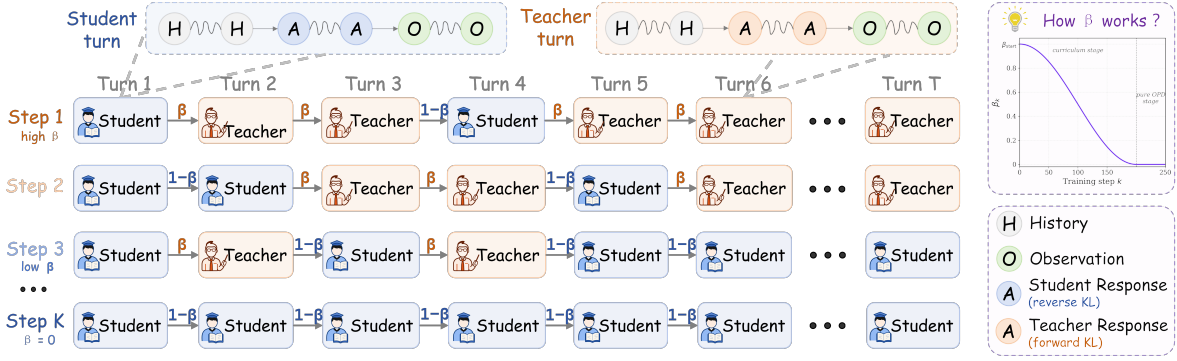}
    \caption{Overview of Guided-OPD. The central grid traces a rollout across training as a sequence of turns. Within each turn, a response $A$ is generated token by token from the history $H$ and triggers a new observation $O$, sampled from either the student (reverse-KL target, blue) or the teacher (forward-KL target, orange). The role of every turn is drawn independently at probability $\beta$ for the teacher and $1-\beta$ for the student, where $\beta$ is set by the curriculum schedule shown in the top-right inset and decays smoothly from $\beta_{\text{start}}$ to zero (cosine by default). Early in training (Step 1, high $\beta$), most turns are produced by the teacher and the rollout is anchored within the teacher's familiar distribution. As training proceeds (Step 2 and Step 3), $\beta$ decreases and the share of student turns grows. Once $\beta$ reaches zero in the pure-OPD stage (Step $K$), every turn is generated by the student and the procedure coincides with standard OPD.}
    \label{fig:method}
\end{figure*}

In this section, we propose Guided-OPD, an on-policy distillation algorithm designed for multi-turn agents and illustrated in Figure~\ref{fig:method}. Its central idea is to inject teacher guidance during the student rollout at the granularity of a single turn and to schedule this guidance in a curriculum fashion. 
Early in training, the teacher takes over a complete turn with relatively high probability, pulling the trajectory back toward the state distribution familiar to the teacher and alleviating the supervision degradation that vanilla OPD suffers from when errors accumulate over the long-horizon interaction of a multi-turn agent. 
As training progresses, the probability of teacher intervention decays along a predefined schedule and eventually reaches zero, so the training process converges smoothly to a pure OPD regime that is fully consistent with inference. Concretely, Guided-OPD consists of two complementary components. 
Turn-level Guidance (\S\ref{sec:method:turn}) specifies the granularity of teacher intervention and the distillation loss assigned to turns generated by each role, while Curriculum Guidance (\S\ref{sec:method:curriculum}) specifies how the teacher intervention probability $\beta$ decays with the training step. The unified optimization objective is then presented in \S\ref{sec:method:objective}.

\subsection{Turn-level Guidance}
\label{sec:method:turn}

\paragraph{The turn as the unit of guidance.}
We refer to the complete ReAct~\citep{react} style thought-plus-action response $a_t$ that the agent produces conditioned on history $h_t$, excluding the subsequent environment observation $o_{t+1}$, as a turn. We choose the turn rather than the token as the unit of guidance because, in a multi-turn agent, the quantity that actually shifts the subsequent state distribution is not an individual token but a complete action together with the observation it triggers~\citep{zhong2026sod}. Switching between teacher and student at turn boundaries therefore aligns the role transitions with the underlying state transitions of the environment.

\paragraph{Role assignment within a turn.}
At this granularity, we draw on the stochastic mixing idea from imitation learning~\citep{DAgger} and introduce a mixing parameter $\beta \in [0, 1]$ that controls the strength of teacher intervention during rollout. Independently at each turn of every trajectory, $a_t$ is generated by the teacher with probability $\beta$ and by the student with probability $1 - \beta$,
\begin{equation}
    a_t \sim
    \begin{cases}
        \pi_\theta(\cdot \mid h_t), & \text{w.p. } 1 - \beta \quad \text{(student turn)}, \\
        \pi_\phi(\cdot \mid h_t),   & \text{w.p. } \beta \quad \text{(teacher turn)}.
    \end{cases}
    \label{eq:turn-mixing}
\end{equation}
The value of $\beta$ is supplied by the curriculum schedule of \S\ref{sec:method:curriculum} according to the current training progress and is held fixed within a single trajectory. Regardless of which role produces $a_t$, the response is appended to the shared interaction history, so that the next-turn history $h_{t+1}$ retains an identical prompt structure under both roles.

\paragraph{Loss on a student turn.}
When a turn is produced by the student, we minimize the reverse KL between the student and the teacher on that turn,
\begin{equation}
    \mathcal{L}_{\text{stu}}^{(t)}(\theta)
    = \mathcal{D}_{\mathrm{KL}}\!\big(\pi_\theta(\cdot \mid h_t) \,\Vert\, \pi_\phi(\cdot \mid h_t)\big).
    \label{eq:student-loss}
\end{equation}

\paragraph{Loss on a teacher turn.}
When a turn is produced by the teacher, we minimize the forward KL between the teacher and the student on that turn,
\begin{equation}
    \mathcal{L}_{\text{tea}}^{(t)}(\theta)
    = \mathcal{D}_{\mathrm{KL}}\!\big(\pi_\phi(\cdot \mid h_t) \,\Vert\, \pi_\theta(\cdot \mid h_t)\big).
    \label{eq:teacher-loss}
\end{equation}

\paragraph{Why this asymmetry?}
The choice of divergence is not arbitrary; it is matched to the role that drives generation on each turn.

On a student turn, the student is the party actually advancing the trajectory, and what we care about is whether every action the student produces is one that the teacher would endorse. The reverse KL is mode-seeking~\citep{opd, opdsurvey}: it requires the actions sampled by the student to lie in high-probability regions of the teacher, while not requiring the student to cover every action the teacher would support. This concentrates the training signal on correcting what the student currently outputs, tightening the student on its own turns toward the modes preferred by the teacher.

On a teacher turn, the teacher is demonstrating an alternative way of behaving in that state, and we want the student to absorb the full demonstration into its own repertoire. The forward KL is mass-covering: it penalizes the student for placing too little probability on the actions that the teacher actually produces, requiring the support of the student to cover all behaviors the teacher chooses in that state. Applying reverse KL on a teacher turn would instead encourage the student to evaluate the demonstration through the lens of its current preferences and to discard those actions that the student has not yet learned to appreciate but that the teacher considers reasonable, which contradicts the very purpose of introducing demonstrations.

In short, the asymmetry aligns the divergence with the role of the turn: a student turn uses reverse KL to tighten the student output onto modes preferred by the teacher, while a teacher turn uses forward KL to make the student cover all reasonable behaviors demonstrated by the teacher.

\subsection{Curriculum Guidance}
\label{sec:method:curriculum}

In the previous subsection, $\beta$ governs the probability that any given turn is produced by the teacher or by the student. We now specify how $\beta$ varies with training progress, so that the strong teacher guidance applied at the start of training gradually fades and the procedure eventually reverts to pure OPD.

\paragraph{Curriculum stage.}
Out of the total $T_{\text{total}}$ training steps, we reserve the first fraction $\rho$ (the curriculum ratio, with default $\rho = 0.8$) as the curriculum stage, comprising $T_{\text{decay}} := \rho T_{\text{total}}$ steps. During this stage, $\beta$ decays smoothly from its initial value $\beta_{\text{start}}$ along a predefined curve to zero. Over the remaining steps, $\beta$ is held at zero, so that the tail of training is strictly equivalent to standard OPD.

\paragraph{Decay schedules.}
Let the decay progress be $\xi_t = \min(t / T_{\text{decay}}, 1)$. We consider three candidate schedules,
\begin{align}
\text{Linear:}\quad
\beta_t
&= \beta_{\mathrm{start}}
 + (\beta_{\mathrm{end}} - \beta_{\mathrm{start}})\xi_t,
\label{eq:beta-linear}\\
\text{Cosine:}\quad
\beta_t
&= \beta_{\mathrm{start}}
 + (\beta_{\mathrm{end}} - \beta_{\mathrm{start}}) \notag\\
&\quad {}\cdot
\frac{1 - \cos(\pi \xi_t)}{2},
\label{eq:beta-cosine}\\
\text{Exp:}\quad
\beta_t
&= \beta_{\mathrm{start}}
\left(
\frac{\beta_{\mathrm{end}}}{\beta_{\mathrm{start}}}
\right)^{\xi_t},
\label{eq:beta-exp}\\[-0.3em]
&\quad \beta_{\mathrm{start}}, \beta_{\mathrm{end}} > 0.
\nonumber
\end{align}
All three schedules satisfy the boundary conditions $\beta_0 = \beta_{\text{start}}$ and $\beta_{T_{\text{decay}}} = \beta_{\text{end}}$, and they differ only in their curvature inside the curriculum. 
The linear schedule decays at a constant rate, and the exponential schedule decays multiplicatively. 
We adopt the cosine schedule by default because the cosine schedule changes slowly near both endpoints and faster in between, which matches the intuition of retaining strong guidance early on while exiting gently near the end.

\subsection{Final Objective}
\label{sec:method:objective}

Combining the per-turn loss of \S\ref{sec:method:turn} with the $\beta$ schedule of \S\ref{sec:method:curriculum} gives the full Guided-OPD objective. At training step $t$, we sample a trajectory $\tau$ from the turn-level mixed rollout distribution $\pi_{\theta, \phi}^{\beta_t}$ induced by Eq.~\ref{eq:turn-mixing} and accumulate the corresponding per-turn losses,
\begin{equation}
\begin{aligned}
&\mathcal{L}_{\text{Guided-OPD}}(\theta; t)
= \mathbb{E}_{\tau \sim \pi_{\theta,\phi}^{\beta_t}}
\!\Bigg[ \\[-0.25em]
&\sum_{u=0}^{T-1}
\Big(
(1-z_u)\mathcal{L}_{\text{stu}}^{(u)}(\theta)
+ z_u\mathcal{L}_{\text{tea}}^{(u)}(\theta)
\Big)
\Bigg].
\end{aligned}
\label{eq:guided-opd-objective}
\end{equation}
where the role indicator $z_u \in \{0, 1\}$ is sampled together with $a_u$ according to Eq.~\ref{eq:turn-mixing}, and the two per-turn losses $\mathcal{L}_{\text{stu}}^{(u)}$ and $\mathcal{L}_{\text{tea}}^{(u)}$ are defined in Eq.~\ref{eq:student-loss} and Eq.~\ref{eq:teacher-loss}.

Taking expectations turn by turn, Eq.~\ref{eq:guided-opd-objective} admits the equivalent form
{\small
\begin{equation}
    \mathcal{L}_{\text{Guided-OPD}}(\theta;t)
    = (1-\beta_t)\mathcal{J}_{\mathrm{rKL}}^{\pi_\theta}(\theta)
    + \beta_t\mathcal{J}_{\mathrm{fKL}}^{\pi_\phi}(\theta).
    \label{eq:guided-opd-expectation}
\end{equation}
}
which decomposes Guided-OPD into reverse-KL distillation on the student-induced state distribution and forward-KL distillation on the teacher-induced state distribution, weighted by $1-\beta_t$ and $\beta_t$ respectively. As $\beta_t \to 0$, the objective recovers standard OPD (Eq.~\ref{eq:opd}) and is fully consistent with the student-only rollout used at inference. As $\beta_t \to 1$, it reduces to forward-KL distillation on teacher trajectories, a safe imitation regime that anchors the student in the teacher's state distribution. The curriculum schedule starts training from the latter regime and transitions smoothly to the former, jointly addressing the trajectory drift and the training-inference mismatch identified in \S\ref{sec:intro}. The complete procedure is summarized in Algorithm~\ref{alg:guided-opd}.

\begin{algorithm}[t]
\small
\DontPrintSemicolon
\caption{Guided-OPD step}
\label{alg:guided-opd}

\KwIn{Teacher $\pi_\phi$, student $\pi_\theta$, environment $\mathcal{E}$, step $t$.}
\KwParam{$\rho$, $T_{\mathrm{total}}$, $\beta_{\mathrm{start}}$, schedule $\mathcal{S}$.}

$T_{\mathrm{decay}} \leftarrow \rho T_{\mathrm{total}}$\;
$\beta_t \leftarrow \mathcal{S}(t,T_{\mathrm{decay}})$
\tcp*{default cosine}

\BlankLine
\textbf{Mixed rollout.}\;
$h_0 \leftarrow \mathcal{E}.\textsc{Reset}()$;
$\tau \leftarrow [\,]$\;

\For{$u=0,1,\ldots$ until terminal}{
  Sample $z_u$ with $\Pr(z_u=\mathrm{tea})=\beta_t$\;

  \eIf{$z_u=\mathrm{tea}$}{
    $a_u \sim \pi_\phi(\cdot \mid h_u)$\;
  }{
    $a_u \sim \pi_\theta(\cdot \mid h_u)$\;
  }

  $o_{u+1} \leftarrow \mathcal{E}.\textsc{Step}(a_u)$\;
  $h_{u+1} \leftarrow (h_u,a_u,o_{u+1})$\;
  Append $(h_u,a_u,z_u)$ to $\tau$\;
}

\BlankLine
\textbf{Asymmetric distillation.}\;
$\mathcal{L}_{\mathrm{stu}}
\leftarrow
\sum_{u:\,z_u=\mathrm{stu}}
\mathcal{L}_{\mathrm{stu}}^{(u)}(\theta)$\;

$\mathcal{L}_{\mathrm{tea}}
\leftarrow
\sum_{u:\,z_u=\mathrm{tea}}
\mathcal{L}_{\mathrm{tea}}^{(u)}(\theta)$\;

$\mathcal{L}(\theta)
\leftarrow
\mathcal{L}_{\mathrm{stu}}+\mathcal{L}_{\mathrm{tea}}$
\tcp*{Eqs.~\eqref{eq:student-loss}, \eqref{eq:teacher-loss}}

\BlankLine
\textbf{Optimizer step.}\;
$\theta \leftarrow \theta - \eta \nabla_\theta \mathcal{L}(\theta)$\;

\end{algorithm}

\section{Experiments}
\label{sec:exp}

\subsection{Experimental Setup}
\label{sec:exp:setup}

\paragraph{Implementation.}
We distill three Qwen3 students of size $0.6$B, $1.7$B, and $4$B from the same teacher Qwen3-30B-A3B~\citep{qwen2025qwen3}. For Guided-OPD, we use a cosine $\beta$ schedule (Eq.~\ref{eq:beta-cosine}) with $\beta_{\text{start}}=1$, $\beta_{\text{end}}=0$, and curriculum ratio $\rho=0.8$. All methods are implemented on Trinity-RFT~\citep{pan2025trinity} and run on $8\times$ NVIDIA H20 GPUs. Full hyperparameters are deferred to Appendix~\ref{app:hparam}.

\paragraph{Baselines.}
We compare Guided-OPD against four baselines spanning training-free evaluation, standard OPD, teacher guidance, and curriculum learning: (i) \emph{Zero-shot}, the untrained student; (ii) \emph{Vanilla OPD}~\citep{opd, lu2025onpolicydistillation}, the standard on-policy distillation of \S\ref{sec:preliminary}; (iii) \emph{AdaSwitch}~\citep{AdaSwitch}, which hands the rest of a generation to the teacher the first time the token-level divergence exceeds an adaptive threshold; and (iv) \emph{TCOD}~\citep{wang2026tcod}, a curriculum method that linearly grows the number of turns $k$ participating in the KL computation. Full implementation details are in Appendix~\ref{app:baselines}.

\paragraph{Evaluation.}
We evaluate on three multi-turn agent benchmarks covering embodied navigation, scientific reasoning, and e-commerce: ALFWorld~\citep{ALFWorld}, ScienceWorld~\citep{ScienceWorld}, and WebShop~\citep{WebShop}; for ALFWorld we use its official \emph{seen}/\emph{unseen} splits as IID and OOD evaluations. We report three metrics: \emph{Score}\,$\uparrow$ (continuous task score in $[0,100]$, unavailable on ALFWorld), \emph{Success Rate} (SR\,$\uparrow$, binary completion rate), and \emph{Round}\,$\downarrow$ (average interaction steps per task). 
All methods are evaluated on the validation set of each benchmark under the same decoding configuration. 
See Appendix~\ref{app:bench} for benchmark details (including Table~\ref{tab:dataset_summary}) and Appendix~\ref{app:metrics} for metric definitions.

\subsection{Main Results}
\label{sec:exp:main}

\begin{table*}[t]
\centering
\small
\setlength{\tabcolsep}{3.2pt}
\renewcommand{\arraystretch}{1.05}
\caption{Main results on three multi-turn agent benchmarks (\textsc{ALFWorld}, \textsc{ScienceWorld}, \textsc{WebShop}). For \textsc{ALFWorld} we report Success Rate (SR\,$\uparrow$) and average interaction Round (Round\,$\downarrow$) under both IID and OOD evaluation settings. For \textsc{ScienceWorld} and \textsc{WebShop} we additionally report task Score\,$\uparrow$. The teacher (Qwen3-30B-A3B) is shown for reference. Within each student-model group, the best result is highlighted in \textbf{bold}.}
\label{tab:main_results}
\resizebox{\textwidth}{!}{%
\begin{tabular}{ll cccc ccc ccc ccc}
\toprule
\multirow{3}{*}{\textbf{Params}} & \multirow{3}{*}{\textbf{Method}}
 & \multicolumn{4}{c}{\textbf{ALFWorld}}
 & \multicolumn{3}{c}{\textbf{ScienceWorld}}
 & \multicolumn{3}{c}{\textbf{WebShop}}
 & \multicolumn{3}{c}{\textbf{Avg}} \\
\cmidrule(lr){3-6} \cmidrule(lr){7-9} \cmidrule(lr){10-12} \cmidrule(lr){13-15}
 & & \multicolumn{2}{c}{IID} & \multicolumn{2}{c}{OOD}
 & \multirow{2}{*}{Score$\uparrow$} & \multirow{2}{*}{SR$\uparrow$} & \multirow{2}{*}{Round$\downarrow$}
 & \multirow{2}{*}{Score$\uparrow$} & \multirow{2}{*}{SR$\uparrow$} & \multirow{2}{*}{Round$\downarrow$}
 & \multirow{2}{*}{Score$\uparrow$} & \multirow{2}{*}{SR$\uparrow$} & \multirow{2}{*}{Round$\downarrow$} \\
\cmidrule(lr){3-4} \cmidrule(lr){5-6}
 & & SR$\uparrow$ & Round$\downarrow$ & SR$\uparrow$ & Round$\downarrow$ & & & & & & & & & \\
\midrule
\rowcolor{yellow!18}
\multicolumn{15}{c}{\textit{Teacher Models from the Qwen3 Series}} \\
\midrule
30B-A3B & Zero-Shot
 & 43.57 & 22.46 & 40.80 & 22.78
 & 27.73 & 7.25 & 21.62
 & 55.07 & 19.25 & 4.96
 & 41.40 & 27.72 & 17.96 \\
\midrule
\rowcolor{green!15}
\multicolumn{15}{c}{\textit{Student Models from the Qwen3 Series}} \\
\midrule
\multirow{5}{*}{0.6B}
 & Zero-shot     &  0.24 & 30.00 &  0.25 & 29.75 &  5.19 & 2.25 & \textbf{20.26} & 10.37 & 3.75 & 14.03 &  7.78 &  1.62 & 23.51 \\
 & Vanilla OPD   & 19.29 & 26.24 & 16.42 & 26.67 & 10.42 & 1.50 & 26.87 & 39.89 & 8.00 &  6.95 & 25.16 & 11.30 & 21.68 \\
 & AdaSwitch     & 25.11 & 24.93 & 17.91 & 26.27 &  9.56 & 1.25 & 29.43 & 43.41 & 7.75 &  5.08 & 26.49 & 13.01 & 21.43 \\
 & TCOD          & 24.29 & 25.32 & 14.93 & 27.14 & 15.71 & 3.00 & 27.48 & 46.37 & 8.25 & \textbf{3.83} & 31.04 & 12.62 & 20.94 \\
 & \textbf{Guided-OPD} & \textbf{27.14} & \textbf{24.50} & \textbf{19.40} & \textbf{25.90} & \textbf{19.63} & \textbf{3.75} & 24.63 & \textbf{48.44} & \textbf{9.75} & 4.80 & \textbf{34.04} & \textbf{15.01} & \textbf{19.96} \\
\midrule
\multirow{5}{*}{1.7B}
 & Zero-shot     &  8.10 & 28.35 &  7.10 & 27.93 &  6.48 & 0.75 & 24.10 & 28.71 & 10.25 &  9.11 & 17.60 &  6.55 & 22.37 \\
 & Vanilla OPD   & 23.57 & 25.45 & 26.87 & 24.54 & 16.94 & 3.00 & 25.27 & 47.59 & 11.00 &  5.54 & 32.27 & 16.11 & 20.20 \\
 & AdaSwitch     & 25.71 & 24.83 & 23.13 & 25.26 & 17.02 & 2.75 & 26.45 & 46.79 & 10.50 &  5.81 & 31.91 & 15.52 & 20.59 \\
 & TCOD          & 27.86 & 23.87 & 24.63 & 25.01 & 19.08 & 3.25 & 25.26 & 49.47 & \textbf{14.25} & 4.39 & 34.28 & 17.50 & 19.63 \\
 & \textbf{Guided-OPD} & \textbf{32.86} & \textbf{23.34} & \textbf{29.85} & \textbf{23.60} & \textbf{23.24} & \textbf{5.25} & \textbf{23.84} & \textbf{52.63} & 11.75 & \textbf{4.14} & \textbf{37.94} & \textbf{19.93} & \textbf{18.73} \\
\midrule
\multirow{5}{*}{4B}
 & Zero-shot     & 28.57 & 24.25 & 27.86 & 24.73 & 16.21 & 3.75 & \textbf{15.97} & 20.59 &  8.25 & 12.23 & 18.40 & 17.11 & 19.30 \\
 & Vanilla OPD   & 31.43 & 23.36 & 32.09 & 23.34 & 26.45 & 7.50 & 23.88 & 50.87 & 13.50 &  4.53 & 38.66 & 21.13 & 18.78 \\
 & AdaSwitch     & 29.29 & 23.59 & 29.10 & 23.96 & 25.24 & 6.25 & 24.48 & 48.91 & 12.25 &  5.57 & 37.08 & 19.22 & 19.40 \\
 & TCOD          & 32.86 & 23.18 & 31.34 & 24.04 & 25.86 & 9.25 & 23.36 & 53.13 & \textbf{19.00} & \textbf{4.26} & 39.50 & 23.11 & 18.71 \\
 & \textbf{Guided-OPD} & \textbf{37.14} & \textbf{22.76} & \textbf{35.08} & \textbf{23.15} & \textbf{29.50} & \textbf{11.50} & 22.86 & \textbf{55.78} & 17.75 & 4.28 & \textbf{42.64} & \textbf{25.37} & \textbf{18.26} \\
\bottomrule
\end{tabular}%
}
\end{table*}

Table~\ref{tab:main_results} reports the full results on ALFWorld, ScienceWorld, and WebShop. Across the 0.6B, 1.7B, and 4B students, Guided-OPD strictly outperforms all baselines on the three aggregate metrics (average Score, SR, Round) within every student-model group and is best on the vast majority of fine-grained metrics. Against Vanilla OPD, the most direct counterpart, it improves average Score by $+35.3\%$, $+17.6\%$, and $+10.3\%$, and average SR by $+32.8\%$, $+23.7\%$, and $+20.1\%$ on the 0.6B, 1.7B, and 4B students, while reducing average Round by $7.9\%$, $7.3\%$, and $2.8\%$. The simultaneous gain in task quality, success, and interaction efficiency indicates that turn-level guidance combined with a curriculum schedule both corrects long-horizon trajectory drift and yields a more compact policy.

Neither teacher guidance nor curriculum learning alone is sufficient. AdaSwitch provides guidance, but once its switch is triggered it hands the remainder of the generation to the teacher and never returns to the student: the tail of the trajectory is always produced by the teacher during training, yet has to be completed by the student alone at inference. This training-inference mismatch at the trajectory tail leaves AdaSwitch above Vanilla OPD only on 0.6B ($26.49$ vs.\ $25.16$ in average Score), and below it on both 1.7B ($31.91$ vs.\ $32.27$) and 4B ($37.08$ vs.\ $38.66$). 
TCOD provides a curriculum but still samples from the student's own rollouts and therefore cannot repair a state distribution that has already drifted; its average Score gain over Vanilla OPD shrinks rapidly as the student grows ($+23.4\%$, $+6.2\%$, and $+2.2\%$ on 0.6B, 1.7B, and 4B). Guided-OPD combines the strengths of both and consequently attains a higher average Score and SR than either at every student size.

The relative gain of Guided-OPD grows as the student shrinks, from $+10.3\%$ in Score on the 4B student to $+35.3\%$ on the 0.6B student. This matches the drift analysis of Section~\ref{sec:intro}: smaller students push trajectories further out of the teacher's familiar distribution, so turn-level anchoring brings a larger marginal benefit. Notably, the 4B student even surpasses the $7.5\times$ larger 30B-A3B teacher in average Score ($42.64$ vs.\ $41.40$), and independently on ScienceWorld ($29.50$ vs.\ $27.73$) and WebShop ($55.78$ vs.\ $55.07$). On the ALFWorld OOD split, Guided-OPD also attains the best SR at every student size; although its IID--OOD gap is not always the smallest, the consistently stronger OOD SR shows that teacher guidance helps the student inherit the teacher's decision structure without sacrificing generalization to unseen layouts and object combinations.

\subsection{Ablation Study}
\label{sec:exp:ablation}

\begin{table}[t]
\centering
\small
\setlength{\tabcolsep}{4pt}
\renewcommand{\arraystretch}{1.05}
\caption{Ablation study on Qwen3-1.7B distilled from Qwen3-30B-A3B. All variants share the same training configuration as Guided-OPD; the three blocks ablate the asymmetric loss design, the curriculum schedule, and the decay shape, respectively. Score, SR, and Round are averaged across ALFWorld, ScienceWorld, and WebShop following the protocol of Table~\ref{tab:main_results}.}
\label{tab:ablation}
\begin{tabular}{lccc}
\toprule
\textbf{Variant} & Score\,$\uparrow$ & SR\,$\uparrow$ & Round\,$\downarrow$ \\
\midrule
\textbf{Guided-OPD (Ours)} & \textbf{37.94} & \textbf{19.93} & \textbf{18.73} \\
\midrule
\multicolumn{4}{l}{(a) Asymmetric loss design} \\
\quad fKL on stu, rKL on tea & 33.32 & 15.47 & 23.30 \\
\quad rKL on both turns      & 36.58 & 17.83 & 20.77 \\
\quad fKL on both turns      & 34.56 & 17.02 & 21.13 \\
\midrule
\multicolumn{4}{l}{(b) Curriculum vs fixed mixing} \\
\quad Fixed $\beta = 0.25$   & 34.41 & 16.09 & 19.97 \\
\quad Fixed $\beta = 0.50$   & 35.17 & 15.62 & 20.05 \\
\quad Fixed $\beta = 0.75$   & 35.85 & 14.82 & 19.36 \\
\midrule
\multicolumn{4}{l}{(c) Decay shape} \\
\quad Linear                 & 36.12 & 17.01 & 20.45 \\
\quad Exp                    & 34.44 & 15.79 & 19.58 \\
\bottomrule
\end{tabular}
\end{table}

To quantify the contribution of each design choice in Guided-OPD, we ablate the three core components introduced in \S\ref{sec:method} on the Qwen3-1.7B student, keeping the teacher, benchmarks, hyperparameters, and evaluation script identical to the main experiments. The three blocks of Table~\ref{tab:ablation} mirror the bottom-up construction of the method: block (a) ablates the asymmetric loss assignment from Turn-level Guidance (\S\ref{sec:method:turn}), block (b) ablates whether $\beta$ must follow a curriculum at all (\S\ref{sec:method:curriculum}), and block (c) ablates the shape of the decay curve. We discuss the three blocks in turn.

\paragraph{Reverse and forward KL are not interchangeable.}
We first examine the asymmetric loss assignment within a turn (Table~\ref{tab:ablation}(a)). Swapping the two KL roles---forward KL on student turns and reverse KL on teacher turns---drops Score and SR sharply from $37.94 / 19.93$ to $33.32 / 15.47$ and raises Round from $18.73$ to $23.30$, the worst Score and Round across all variants. This swap misuses both divergences at once: forward KL on student turns asks the student to mass-cover the teacher while it is actively driving the rollout, encouraging it to emit unfamiliar teacher actions and accelerating the trajectory drift identified in \S\ref{sec:intro}; reverse KL on teacher turns lets the student re-weight the demonstration through its current preferences, partially discarding the high-quality behaviors it is meant to inject. Reverse KL on both turns ($36.58$ Score, $17.83$ SR) clearly beats forward KL on both ($34.56$, $17.02$), confirming that tightening the student with mode-seeking reverse KL on its own trajectories matters more than forcing it to cover the full teacher behavior. The default asymmetric pairing outperforms both symmetric variants, showing that matching each divergence to the role that drives generation is what makes the loss design effective.

\paragraph{The curriculum is not replaceable by fixed mixing.}
We then examine whether $\beta$ needs to decay with training. Table~\ref{tab:ablation}(b) shows that all fixed-$\beta$ variants trail the cosine schedule (Score $37.94$) by $2.1$--$3.5$ points, with an even larger SR gap: the best, fixed-$\beta=0.25$, reaches only $16.09$, $3.84$ points below our $19.93$. Strikingly, as fixed $\beta$ grows from $0.25$ to $0.75$, Score keeps rising ($34.41 \to 35.17 \to 35.85$) while SR moves in the opposite direction ($16.09 \to 15.62 \to 14.82$). More teacher intervention is thus not always better: a persistent teacher exposure trains the student into an imitator that reproduces intermediate teacher behaviors---scoring higher under the partial-credit Score---but never operates on its own rollouts long enough to learn how to recover without a teacher hint, which costs it on the stricter binary SR and is mirrored by a Round that stays near $20$ for every fixed $\beta$. The cosine curriculum resolves this dilemma by concentrating guidance in the early stage when the student is weakest and OOD drift most likely, and gradually withdrawing it as the student matures, achieving the best Score, SR, and Round simultaneously.

\paragraph{The cosine shape strikes the best trade-off.}
Finally, we compare the three decay shapes while keeping all other configurations fixed. As shown in Table~\ref{tab:ablation}(c), cosine ($37.94$ Score) strictly outperforms linear ($36.12$) and exp ($34.44$), with exp suffering the largest drop on both Score and SR. The cosine curve decays slowly near both endpoints and faster in the middle, matching the intuition of \S\ref{sec:method:curriculum}: retain a strong teacher anchor while the student is still fragile, transfer the bulk of supervision in the middle phase, and exit smoothly so that late training is indistinguishable from pure OPD. The linear schedule preserves moderate guidance throughout and is a reasonable but suboptimal backup. 
The exponential schedule, in contrast, decays $\beta$ too aggressively in the early stage and removes teacher support precisely at the phase where OOD errors accumulate most easily, which explains its largest drop.

\section{Related Work}
\label{sec:related}

\paragraph{On-Policy Distillation for Multi-turn Agents.}
Standard OPD~\citep{opd, lu2025onpolicydistillation} excels on single-turn reasoning~\citep{eopd, tip, yang2026learning} but suffers in multi-turn settings as early errors compound
and degrade subsequent supervision~\citep{DAgger, fu2026revisiting, li2026rethinking}; 
recent work either augments the token-level KL with reward-aligned signals from verifiable rewards or reward models~\citep{zhong2026sod, skillsd, srpo}, which depends on a reliable verifier or critic~\citep{deepseek2025r1, grpo, yu2025dapo}, 
or schedules trajectory depth via a temporal curriculum~\citep{wang2026tcod} that operates only at the macro level and does not repair the state distribution along trajectories the student has already entered.
We instead intervene inside the rollout through explicit teacher guidance~\citep{DAgger}, pulling the state distribution back into the teacher-familiar region without any extra reward or critic.

\paragraph{Teacher Guidance during Distillation Rollout.}
A more direct line of work injects teacher guidance into the rollout itself~\citep{SKD, AdaSwitch}, but existing methods sit at two unbalanced extremes: token-level bidirectional switching~\citep{SWITCH, SKD} alternates at every token and produces chimeric trajectories~\citep{AdaSwitch}, while sequence-tail one-shot handover~\citep{AdaSwitch} transfers the remainder of generation to the teacher upon a single deviation and prevents the student from recovering from local errors at later decoding steps.
Both keep the teacher present throughout training and leave a persistent training-inference mismatch when the teacher is removed at inference~\citep{fu2026revisiting}.
Guided-OPD intervenes at the turn level, the granularity aligned with the environment's state transitions~\citep{xue2025simpletir,zhou2024archer}, and decays this intervention to zero by the end of training, so that the late stage returns to pure OPD~\citep{opd} and the mismatch is eliminated.

\section{Conclusion}
We studied On-Policy Distillation in the multi-turn agent setting and identified a characteristic failure mode in which small student errors compound across turns and push the rollout out of the teacher's familiar state distribution, so that the teacher's supervision becomes least reliable where the student needs it most. We addressed this with Guided-OPD, a simple yet effective algorithm that mixes teacher- and student-generated turns within a rollout and schedules the teacher's intervention along a curriculum decaying smoothly to zero. Turn-level mixing aligns role transitions with environment state transitions, the asymmetric pairing of reverse and forward KL matches each divergence to the role driving generation, and the curriculum preserves the training-inference consistency. Across ALFWorld, ScienceWorld, and WebShop, and across Qwen3 students of three scales distilled from Qwen3-30B-A3B, Guided-OPD consistently outperforms vanilla OPD and strong teacher-guidance and curriculum baselines, with the largest gains on the smallest student. We hope the rollout-drift perspective and turn-level curriculum view will inform future work on transferring agentic capabilities to compact, deployable models.

\section*{Limitations}
Our study has several limitations that point to natural directions for future work.
First, the evaluation suite covers only three multi-turn agent benchmarks, namely ALFWorld, ScienceWorld, and WebShop, all of which expose the agent through purely text-based observations and a discrete action set. Although these environments already span embodied navigation, scientific reasoning, and e-commerce, they leave out a number of regimes that are increasingly important for practical agents, including tool-use agents with open-ended API calls, software engineering agents that operate on real codebases, GUI agents that perceive screenshots, and embodied agents that act in continuous physical environments. Whether Guided-OPD remains effective under these richer observation and action spaces is an open question.
Second, all distillation experiments are conducted within a single model family. We distill Qwen3 students of three different scales from the same Qwen3-30B-A3B teacher, so the conclusions reflect intra-family transfer with aligned tokenizers and similar inductive biases. Cross-family distillation, for example from a Qwen teacher to a Llama or Mistral student, may introduce tokenizer mismatches and distributional gaps that further amplify the rollout-drift problem and possibly require adjustments to the curriculum schedule; combining Guided-OPD with recent cross-tokenizer distillation techniques~\citep{sun2026simct} is a promising direction.
Third, the curriculum hyperparameters, including the initial mixing probability $\beta_{\text{start}}$, the curriculum ratio $\rho$, and the cosine decay shape, are chosen once and shared across all student sizes and benchmarks. In contrast, a schedule that adapts $\beta$ to a measured training signal rather than following a predefined curve may reduce this tuning cost and raise the performance ceiling.
Finally, our experiments are run on a single 8-GPU node with students up to 4B parameters and a 30B-A3B teacher. Verifying the trends at substantially larger student or teacher scales, and on longer-horizon tasks where rollout drift is likely to be more severe, requires resources beyond those available for this study and is left to future work.



\bibliography{custom}

\appendix

\section{Experimental Setup Details}
\label{app:exp_setup}

This appendix provides the full experimental details that complement \S\ref{sec:exp:setup}. The five subsections describe the benchmark environments (\S\ref{app:bench}), the evaluation metrics (\S\ref{app:metrics}), the baselines (\S\ref{app:baselines}), the training and evaluation hyperparameters (\S\ref{app:hparam}), and the prompt templates used for each environment (\S\ref{app:prompts}).

\subsection{Benchmark Environments}
\label{app:bench}

We evaluate Guided-OPD on three text-based multi-turn interactive environments that cover three complementary task families: embodied navigation, scientific reasoning, and e-commerce. The dataset-level statistics are summarized in Table~\ref{tab:dataset_summary}.

\paragraph{ALFWorld.}
ALFWorld~\citep{ALFWorld} is a text-based embodied environment that requires the agent to complete six categories of household tasks: picking a target object and placing it at a destination; picking and placing after cleaning, heating, or cooling the object; examining an object under a light source; and picking and placing two objects of the same type. The agent perceives the environment through natural-language observations and acts through high-level commands such as moving to a location, opening containers, taking and putting objects, heating, cooling, and examining items, with a maximum of $30$ interaction steps per episode. ALFWorld provides two official evaluation splits: the \emph{seen} split shares room layouts and object combinations with the training distribution and serves as our \textbf{IID} evaluation, while the \emph{unseen} split contains novel room layouts and object combinations never observed during training and serves as our \textbf{OOD} evaluation. The OOD split therefore probes whether the policy learned by Guided-OPD generalizes beyond the specific environments encountered during distillation.

\paragraph{ScienceWorld.}
ScienceWorld~\citep{ScienceWorld} is a text-based interactive environment that tests scientific reasoning across $30$ task types aligned with an elementary-school science curriculum, covering topics such as thermodynamics, electric circuits, chemistry, biology, and matter classification. The environment exposes a large action space (e.g., \texttt{focus on}, \texttt{activate}, \texttt{mix}, \texttt{wait}, \texttt{move \dots to \dots}) together with a structured observation that lists the agent's current location, visible objects, and inventory. Each episode allows at most $30$ interaction steps, and at the end of each episode the environment returns a continuous score in $[0, 100]$ that reflects the fraction of subgoals completed. ScienceWorld is the most challenging benchmark in our suite because successful trajectories typically require long-horizon planning, multi-step tool use, and a non-trivial chain of cause--effect reasoning.

\paragraph{WebShop.}
WebShop~\citep{WebShop} is a simulated e-commerce environment in which the agent must satisfy a natural-language shopping instruction (e.g., specific product type, color, size, and price range) by interacting with a structured web interface. The agent observes a rendered HTML page that includes a search bar, product listings, and product detail pages, and acts through two action primitives: \texttt{search[query]} for issuing search queries and \texttt{click[button\_name]} for navigating links, applying filters, or completing a purchase. Each episode allows at most $15$ interaction steps. After purchase, the environment returns a continuous score in $[0, 100]$ that measures how well the purchased product matches the requested attributes, and a binary success signal when the score reaches $100$.

\begin{table}[t]
\centering
\scriptsize
\setlength{\tabcolsep}{2pt}
\caption{Summary of the benchmarks used. ``Max Turns'' is the maximum number of agent--environment interaction rounds per task.}
\label{tab:dataset_summary}
\begin{tabular}{lccc}
\toprule
\textbf{Benchmark}  & \textbf{Type} & \textbf{Difficulty} & \textbf{Max Turns} \\
\midrule
ALFWorld (IID)        & Embodied             & Easy   & $30$ \\
ALFWorld (OOD)        & Embodied             & Medium & $30$ \\
ScienceWorld          & Scientific Reasoning & Hard   & $30$ \\
WebShop               & E-commerce           & Medium & $15$ \\
\bottomrule
\end{tabular}
\end{table}

\subsection{Metrics}
\label{app:metrics}

We report three metrics in a unified manner across the three benchmarks.

\paragraph{Score ($\uparrow$).}
The continuous task score natively provided by ScienceWorld and WebShop, bounded in $[0, 100]$. On ScienceWorld it is computed from the fraction of subgoals completed during the episode, and on WebShop it is computed from the attribute-level match between the purchased product and the requested specification. Score offers a partial-credit, fine-grained measurement that lets us distinguish between failures that are close to completion and failures that are entirely off the mark. ALFWorld only provides a binary success signal, so Score is not reported on it.

\paragraph{Success Rate (SR\,$\uparrow$).}
The task completion rate, treated as a binary outcome on every task and bounded above by $\mathrm{SR}=100\%$ and below by $0\%$. 
On ALFWorld an episode is counted as successful when the environment emits a terminal success flag; 
on ScienceWorld and WebShop, where an episode may terminate without full credit, an episode is counted as successful only when it receives a Score of $100$.

\paragraph{Round ($\downarrow$).}
The average number of environment interaction steps consumed per task, which reflects the efficiency of the learned policy. 
When SR is similar across methods, a smaller Round is preferred because it indicates that the policy completes the task with fewer interactions.

For ALFWorld we additionally report SR and Round separately on the IID and OOD splits, so that in-distribution performance and OOD generalization are measured at the same time. 
All metrics are averaged over the validation set of each benchmark, and all methods use the same decoding configuration to ensure a fair comparison.

\begin{table*}[t]
\centering
\caption{Training hyperparameters across all environments. The left block lists the configuration shared by all methods; the right block lists the system-level setup and the hyperparameters specific to Guided-OPD and to the two baselines that have method-specific hyperparameters (AdaSwitch, TCOD).}
\setlength{\tabcolsep}{2.8pt}
\label{tab:train_params}
\small
\begin{minipage}[t]{0.48\textwidth}
\centering
\begin{tabular}{ll}
\toprule
\textbf{Hyperparameter} & \textbf{Value} \\
\midrule
\textbf{Algorithm} & \\
\quad Algorithm type & On-Policy Distillation \\
\quad KL coefficient & $1.0$ \\
\quad Learning rate & $1 \times 10^{-6}$ \\
\quad Optimizer & AdamW \\
\quad Gradient clipping & $1.0$ \\
\quad Repeat times & $1$ \\
\quad Sample strategy & Staleness ctrl ($\Delta_{\max}\!=\!2$) \\
\midrule
\textbf{Training} & \\
\quad Total training steps & $250$ \\
\quad Batch size & $16$ \\
\quad Train batch size & $64$ \\
\quad Save interval & $250$ \\
\quad Evaluation interval & $5$ steps \\
\midrule
\textbf{Model Configuration} & \\
\quad Max prompt tokens & $10{,}240$ \\
\quad Max response tokens & $512$ \\
\midrule
\textbf{Inference (Rollout)} & \\
\quad Temperature (training) & $1.0$ \\
\quad Temperature (evaluation) & $0.4$ \\
\quad Logprobs & Enabled (all tokens) \\
\quad Seed & $42$ \\
\midrule
\textbf{Environment-Specific} & \\
\quad ALFWorld max steps & $30$ \\
\quad ScienceWorld max steps & $30$ \\
\quad WebShop max steps & $15$ \\
\bottomrule
\end{tabular}
\end{minipage}\hfill
\begin{minipage}[t]{0.48\textwidth}
\centering
\begin{tabular}{ll}
\toprule
\textbf{Hyperparameter} & \textbf{Value} \\
\midrule
\textbf{Distributed Training} & \\
\quad Number of nodes & $1$ \\
\quad GPUs per node & $8$ (NVIDIA H20, 96\,GB) \\
\quad GPU allocation & $4$ act.\,/\,$2$ tea.\,/\,$2$ lrn. \\
\quad Tensor parallel size & $2$ \\
\quad Sequence parallel size & $2$ (Ulysses) \\
\quad Max tokens per GPU & $16{,}384$ \\
\quad GPU memory utilization & $0.7$ \\
\quad Data type & BFloat16 \\
\midrule
\textbf{Guided-OPD (Ours)} & \\
\quad Init.\ mixing $\beta_{\text{start}}$ & $1.0$ \\
\quad Term.\ mixing $\beta_{\text{end}}$ & $0$ \\
\quad Curriculum ratio $\rho$ & $0.8$ \\
\quad Decay schedule & Cosine (Eq.~\ref{eq:beta-cosine}) \\
\quad Student-turn loss & rKL (Eq.~\ref{eq:student-loss}) \\
\quad Teacher-turn loss & fKL (Eq.~\ref{eq:teacher-loss}) \\
\midrule
\textbf{AdaSwitch~\citep{AdaSwitch}} & \\
\quad Sliding window length $L$ & $10$ \\
\quad Threshold multiplier $K$ & $3$ \\
\midrule
\textbf{TCOD~\citep{wang2026tcod}} & \\
\quad Variant & F2B \\
\quad Starting horizon $k_{\text{start}}$ & $1$ \\
\quad Pacing factor $\eta$ & $2$ \\
\bottomrule
\end{tabular}
\end{minipage}
\end{table*}

\subsection{Baselines}
\label{app:baselines}

We benchmark Guided-OPD against four baselines that span training-free evaluation, the standard OPD paradigm, the teacher-guidance family, and the curriculum-learning family. All baselines share the same teacher (Qwen3-30B-A3B), the same training hyperparameters (Table~\ref{tab:train_params}), and the same evaluation script as Guided-OPD to ensure a fair comparison.

\paragraph{Zero-Shot Student (Lower Bound).}
The untrained student $\pi_\theta$ evaluated directly on the interactive tasks without any task-specific fine-tuning or distillation. This establishes the absolute starting point of the student's capability in each environment, and serves as a sanity-check lower bound for all distillation methods.

\paragraph{Vanilla OPD.}
The standard multi-turn adaptation of recent OPD methods~\citep{opd, lu2025onpolicydistillation}, defined in \S\ref{sec:preliminary}. The student samples the entire trajectory by itself and the teacher KL is applied at every token of the agent-generated segments, with no teacher intervention during rollout. This serves as the direct counterpart against which we measure the effect of introducing turn-level teacher guidance and a curriculum schedule.

\paragraph{AdaSwitch.}
AdaSwitch~\citep{AdaSwitch} is a recent representative of the teacher-guidance family. 
It maintains a sliding-window average $\bar{d}$ of the token-level student--teacher divergence over the most recent $L$ steps and, the first time the instantaneous divergence exceeds the adaptive threshold $\tau = K\bar{d}$, hands the remainder of the current generation to the teacher; the handover is one-shot and is never reverted, so no further divergence checks are performed. 
We follow its recommended settings, with threshold multiplier $K = 3$ and sliding window length $L = 10$. 
The switching is fully triggered by sample-level divergence signals and is independent of training progress, so there is no curriculum tied to the training step.

\paragraph{TCOD.}
TCOD~\citep{wang2026tcod} is a recent curriculum-based method tailored to multi-turn agent OPD. It uses the rollout depth $k$, that is, the maximum number of interaction turns the student is allowed to take before the trajectory is truncated, as a proxy for difficulty, and increases $k$ linearly with the training step $n$ via the fixed pacing schedule $k_n = \min(T_{\max}, k_{\text{start}} + \lfloor n / \eta \rfloor)$, so that the student gradually expands from short-horizon decisions to full trajectories and the KL is accordingly accumulated over these $k$ turns. 
We follow its recommended fixed pacing with $\eta = 2$ and $k_{\text{start}} = 1$, and adopt its F2B variant because it does not require pre-collected successful teacher trajectories. The curriculum acts purely on trajectory depth, and no turn-level teacher intervention is introduced inside the rollout.

\begin{table}[t]
\centering
\caption{Evaluation hyperparameters across all environments.}
\label{tab:eval_params}
\small
\setlength{\tabcolsep}{2.6pt}
\begin{tabular}{ll}
\toprule
\textbf{Hyperparameter} & \textbf{Value} \\
\midrule
\textbf{Generation} & \\
\quad Maximum tokens & $4{,}096$ \\
\quad Temperature & $0.4$ \\
\quad Top-p / Top-k / MinP & $1.0$ / $-1$ / $0.0$ \\
\midrule
\textbf{Environment} & \\
\quad Max env.\ steps & $30$ (ALFW.\,/\,SciW.) \,/\, $15$ (WebS.) \\
\quad History length & $2$ steps \\
\midrule
\textbf{Parallelization} & \\
\quad Number of workers & $8$ \\
\quad Process timeout & $3{,}600$\,s \\
\quad Synchronization style & Dynamic by explorer \\
\midrule
\textbf{Data Splits} & \\
\quad ALFWorld & \texttt{test}\,/\,\texttt{test\_unseen} \\
\quad ScienceWorld & Test split \\
\quad WebShop & Test split \\
\bottomrule
\end{tabular}
\end{table}

\subsection{Hyperparameters}
\label{app:hparam}

We conduct training across the three text-based interactive environments described in \S\ref{app:bench}. The training configuration shared by all methods is summarized in Table~\ref{tab:train_params}, where the last two blocks list the hyperparameters specific to Guided-OPD and to the two baselines that have method-specific hyperparameters (AdaSwitch, TCOD). Evaluation hyperparameters are summarized in Table~\ref{tab:eval_params}.

\subsection{Prompt Templates}
\label{app:prompts}

This subsection lists the prompt templates used for the three environments during both training and evaluation. All prompts follow a consistent structure: task description, observation--action history, current observation, admissible actions, and a thinking/action format requirement. The agent is required to first produce its reasoning inside \texttt{<thought>} tags and then emit an executable command inside \texttt{<action>} tags, so that the reasoning and the environment-bound action are syntactically separable.

\paragraph{ALFWorld.}
ALFWorld is an embodied task that requires the agent to navigate household environments and complete object-manipulation goals. The prompt emphasizes step-by-step reasoning within \texttt{<thought>} tags followed by an executable action within \texttt{<action>} tags selected from a dynamically updated admissible-action set.

\begin{quote}
\small\ttfamily
You are an expert agent operating in the ALFRED Embodied Environment. Your task is to: \{task\_description\}\\[2pt]
Prior to this step, you have already taken \{step\_count\} step(s). Below are the most recent \{history\_length\} observations and the corresponding actions you took: \{action\_history\}\\[2pt]
You are now at step \{current\_step\} and your current observation is: \{current\_observation\}\\[2pt]
Your admissible actions of the current situation are: [\{admissible\_actions\}].\\[6pt]
Now it's your turn to take an action.\\[2pt]
You should first reason step-by-step about the current situation. This reasoning process MUST be enclosed within <thought> tags.\\[2pt]
Once you've finished your reasoning, you should choose an admissible action for the current step and present it within <action> </action> tags.
\end{quote}

\paragraph{ScienceWorld.}
ScienceWorld focuses on scientific reasoning tasks in a text-based laboratory environment. The prompt guides the agent through multi-step experiments that require domain knowledge and procedural reasoning.

\begin{quote}
\small\ttfamily
Your ScienceWorld task is: \{task\_description\}\\[2pt]
Prior to this step, you have already taken \{step\_count\} step(s). Below are the most recent \{history\_length\} observations and the corresponding actions you took: \{action\_history\}\\[2pt]
You are now at step \{current\_step\} and your current observation is: \{current\_observation\}\\[2pt]
Your valid actions of the current situation are: [\{admissible\_actions\}].\\[6pt]
Now it's your turn to take an action.\\[2pt]
You should first reason step-by-step about the current situation. This reasoning process MUST be enclosed within <thought> tags.\\[2pt]
Once you've finished your reasoning, you should choose a valid action for the current step and present it within <action> </action> tags.
\end{quote}

\paragraph{WebShop.}
WebShop presents e-commerce shopping tasks that require the agent to navigate product listings, apply filters, and make purchasing decisions according to natural-language instructions. The prompt emphasizes matching user preferences to available product attributes.

\begin{quote}
\small\ttfamily
You are an expert autonomous agent operating in the WebShop e-commerce environment.\\[2pt]
Your task is to: \{task\_description\}.\\[2pt]
Prior to this step, you have already taken \{step\_count\} step(s). Below are the most recent \{history\_length\} observations and the corresponding actions you took: \{action\_history\}\\[2pt]
You are now at step \{current\_step\} and your current observation is: \{current\_observation\}.\\[2pt]
Your admissible actions of the current situation are:\\
{[}\\
\{available\_actions\}\\
{]}.\\[6pt]
Now it's your turn to take one action for the current step.\\[2pt]
You should first reason step-by-step about the current situation, then think carefully which admissible action best advances the shopping goal. This reasoning process MUST be enclosed within <thought> tags.\\[2pt]
Once you've finished your reasoning, you should choose an admissible action for the current step and present it within <action> </action> tags.
\end{quote}

WebShop uses a specific action format with two primary action types: \texttt{search[<query>]}, which issues a search using a free-text query and is only available when the search bar is present, and \texttt{click[<button\_name>]}, which clicks an interactive element such as a product link, a filter button, or a pagination button. The set of available clicks is regenerated from the current page state at every step.

\end{document}